# Retrieval-Augmented Generation for Generative Artificial Intelligence in Medicine

Rui Yang, MS[1], Yilin Ning, PhD[1], Emilia Keppo[2], Mingxuan Liu, MS[1], Chuan Hong, PhD[3], Danielle S Bitterman, PhD[4], Jasmine Chiat Ling Ong, PharmD[5], Daniel Shu Wei Ting, PhD[1,6,7], Nan Liu, PhD[1,8,9]*

[1] Center for Quantitative Medicine, Duke-NUS Medical School, Singapore, Singapore

[2] Faculty of Arts and Science, University of Toronto, Toronto, ON, Canada

[3] Department of Biostatistics and Bioinformatics, Duke University, Durham, NC, USA

[4] Artificial Intelligence in Medicine Program, Mass General Brigham, Harvard Medical School, Boston, MA, USA

[5] Division of Pharmacy, Singapore General Hospital, Singapore, Singapore

[6] Singapore Eye Research Institute, Singapore National Eye Center, Singapore, Singapore

[7] Byers Eye Institute, Stanford University, Stanford, CA, USA

[8] Program in Health Services and Systems Research, Duke-NUS Medical School, Singapore, Singapore

[9] Institute of Data Science, National University of Singapore, Singapore, Singapore

*Correspondence: Nan Liu, Centre for Quantitative Medicine, Duke-NUS Medical School, 8 College Road, Singapore 169857, Singapore

Email: liu.nan@duke-nus.edu.sg

## Abstract
Generative artificial intelligence (AI) has brought revolutionary innovations in various fields, including medicine. However, it also exhibits limitations. In response, retrieval-augmented generation (RAG) provides a potential solution, enabling models to generate more accurate contents by leveraging the retrieval of external knowledge. With the rapid advancement of generative AI, RAG can pave the way for connecting this transformative technology with medical applications and is expected to bring innovations in equity, reliability, and personalization to health care.

## Main Text

Generative artificial intelligence (AI) has recently attracted widespread attention across various fields, including the GPT[1,2] and LLaMA[3–5] series for text generation; DALL-E[6] for image generation; as well as Sora[7] for video generation. In medicine, generative AI holds tremendous potential for applications in consulting, diagnosis, treatment, management, and education[8,9]. Additionally, the utilization of generative AI could enhance the quality of health services for patients while alleviating the workload for clinicians[9–11].

Despite this, we must consider the inherent limitations of generative AI models, which include susceptibility to biases from pre-training data[12], lack of transparency, the potential to generate incorrect content, difficulty in maintaining up-to-date knowledge, among others[8]. For instance, large language models were shown to generate biased responses by adopting outdated race-based equations to estimate renal function[13]. In the process of image generation, biases related to gender, skin tone, and geo-cultural factors have been observed[14]. Similarly, for downstream tasks such as question answering and text summarization, the generated content is often factually inconsistent and lacks evidence for verification[15]. Moreover, due to their static knowledge and inability to access external data, generative AI models are unable to provide up to date clinical advice for physicians or effective personalized health management for patients[16].

In tackling these challenges, retrieval-augmented generation (RAG) may provide a solution[17,18]. By providing models access to external data, RAG is capable of enhancing the accuracy of generated content. Specifically, a typical RAG framework consists of three parts (Figure 1): indexing, retrieval, and generation. In the indexing stage, external data is split into chunks, encoded into vectors, and stored into a vector database. In the retrieval stage, the user's query is encoded into a vector representation, and then the most relevant information is retrieved through similarity calculations between the query and the information in the vector database. In the generation stage, both the user's query and the retrieved relevant information are prompted to the model to generate content. Compared to fine-tuning a model for a specialized task, RAG has been shown to improve accuracy for knowledge-intensive tasks[19], and is a more flexible paradigm for model updates.

In this comment, we emphasize the significance of RAG in the era of generative AI, particularly its potential applications within medicine and healthcare. We primarily analyze the breakthroughs that RAG could bring to medicine from three perspectives: equity, reliability, and personalization (Figure 2). Additionally, we explore the limitations of RAG in medical application scenarios.

**Promoting health equity in generative AI applications**

**Bias reduction**
The content generated by generative AI models could perpetuate biases inherent in the pre-training data, which are reflected in aspects including demographic characteristics, political ideologies, and sexual orientations[13,14,20]. Such biases can not only lead to unfair diagnoses and treatments, but may also exacerbate health inequalities for particular populations.

RAG is able to obtain information from external knowledge sources, including medical literature, clinical guidelines, and case reports, to optimize the output of generative AI models[18]. By retrieving information specific to certain subpopulations, the model could analyze a patient's condition from multiple perspectives, potentially reducing the risk of bias contained in the generated content. For instance, when targeting different gender groups, RAG could retrieve research findings on their specific physiological patterns, common disease spectra, clinical manifestations, as well as related recommendations on clinical practice[21–23]. Similarly, for different ethnic groups, RAG enables access to research reports involving their genetic, environmental, and lifestyle factors, to understand potential differences in disease incidence rates and unique symptom presentations[24]. Furthermore, for other specific subpopulations (such as different age groups, socioeconomic statuses, etc.), RAG can retrieve more relevant medical evidence to help comprehensively understand their unique health needs[25]. Although RAG currently may not have sufficient high-quality information on underrepresented groups, it still holds the potential to achieve this goal.

**Disparity mitigation**
Health disparities pose additional challenges to the marginalized groups in accessing medical resources and health services, hindering the achievement of fairness. Although generative AI models are trained on extensive data, the pre-training data itself may exhibit imbalances in representing different groups. For example, 92.64% of the pre-training corpus of GPT-3 is derived from English sources, resulting in limited coverage of communities that speak other languages[1]. This skewness undoubtedly makes it difficult to meet the medical needs of underrepresented groups.

Collecting data specific to these underrepresented populations and incorporating it into the RAG system holds the potential to mitigate the disparities in medicine. Specifically, in low-resource regions, the RAG system could leverage knowledge that integrates local medical research literature, clinical guidelines, and practical experiences to provide more relevant diagnostic and treatment advice to local residents[26]. While these regional guidelines may not be digitized, audio and image recognition technologies have the potential to convert this information into digital format, creating region-specific contextual databases[27]. Similarly, by developing high-quality multilingual medical knowledge bases, RAG can play an important role in cross-language information retrieval and knowledge integration, with the potential to eliminate barriers posed by language differences. However, there may be challenges in retrieval effectiveness for minority languages using RAG. Additionally, RAG systems are able to retrieve pre-collected materials and present them in various formats such as text, images, and videos to facilitate patient education. This way allows the explanation of complex medical concepts to patients with different educational and cultural backgrounds[28].

**Generating reliable contents**

**Mistake alleviation**
One significant challenge of generative AI models in medicine is their potential to generate incorrect or unfaithful information[8,9]. Although there are already specific models pre-trained on large amounts of medical data, such as Med-PaLM2 and Med-Gemini, the phenomenon of "hallucination" cannot be avoided[28,29]. This issue is extremely sensitive since any false information related to disease diagnosis, treatment plans, or medication guidance will likely cause serious harm to patients.

For example, medication errors are a major category of medical mistakes, annually resulting in numerous patient fatalities[30,31]. During the stage of converting prescription instructions into a standard format, pharmacy technicians may incorrectly record dosage, frequency, or route of administration[30]. Additionally, when patients transfer medications from their original packaging to other containers, it becomes difficult for pharmacists to recognize the medications, potentially leading to omission errors[31]. Given that electronic health record recommendations and alerts are often imprecise, and traditional NLP methods require extensive human annotation, generative AI could be an attractive solution. However, generative AI models may also generate incorrect drug information, leading to further harm. RAG holds potential to address these issues. By searching various drug information, RAG can automatically parse prescriptions at the data entry stage and generate more accurate medication guidance, thereby reducing medical errors caused by information transmission. Additionally, in the process of drug identification, a multimodal RAG system has the capability to recognize the appearance features of drugs, such as color, shape, and imprints[32], though this technology is still nascent in medicine. By matching these characteristics with database information, the RAG system could generate reliable drug information to serve as a reference for pharmacists, thereby improving the efficiency of drug identification.

**Transparency enhancement**
The "black box" nature of generative AI models makes it difficult to explain how specific diagnoses or treatment recommendations are derived. This lack of transparency not only undermines the trust of physicians and patients in the generated content, but more importantly, it may pose serious medical risks and ethical concerns. Although some research has attempted to enhance models' reasoning abilities and transparency through approaches like chain of thought[33], multi-agent discussion[34], and post-hoc attribution[35], there are still limitations in medical applications[36].

In comparison, RAG is able to retrieve traceable medical facts from external knowledge bases, promoting the generation of more transparent content; however, this process still requires manual verification[37]. In assisting clinical decision-making, RAG can provide the sources of information upon which the diagnoses are based, including clinical guidelines, medical evidence, and clinical cases, ensuring the trustworthiness in the decision support process. Additionally, some research utilizes external medical knowledge graphs (such as the Unified Medical Language System) to enhance the diagnostic capabilities of models. Based on the medical query, the RAG system retrieves knowledge paths directly related to the condition from the graph, such as causes, symptoms, treatments, etc., and leverages this structured knowledge to provide clear diagnostic explanations[15].

**Personalizing healthcare services**

**Health management**
RAG is also showing promise for personalized health care management. Generative AI models lack the ability to incorporate personal information, making it difficult to provide effective health services[9]. For example, they may not be aware of a user's allergies and recommend allergenic foods. In contrast, the RAG system could integrate multimodal health data and lifestyle habits of individuals to build a comprehensive personal profile and provide customized health guidance.

For patients, by connecting their medical records and clinical data while allowing for real-time updates, the RAG system has the capability to provide more precise health management guidance. For instance, for patients with chronic conditions who need to take multiple medications long-term, the system is able to generate medication reminders according to physicians' prescriptions, ensuring that patients take their medications correctly and timely, thereby improving the medication adherence. For the general public, the RAG system can analyze personal health data, lifestyle, environmental factors, and genetic information (if granted access by individual users) to identify potential health risks. In this way, the RAG system provides personalized health recommendations, including diet, exercise, and stress management, effectively promoting disease prevention. For example, for individuals with a high genetic risk of heart disease, the system could recommend specific dietary plans and appropriate exercise regimens to reduce the risk of eventually developing the disease.

**Precision medicine**

Precision medicine aims to maximize medical effectiveness and patient benefits by tailoring treatment strategies according to a patient's genetic profile, environmental influences, lifestyle, and other individual factors[38]. Although current generative AI models have demonstrated potential to assist in clinical decision-making[34,39], they still face challenges in precision medicine[40], as they struggle to utilize highly individualized patient data to provide precise treatment recommendations.

RAG could offer unique advantages for advancing precision medicine. By retrieving a patient's complex clinical and molecular data, the RAG system empowers physicians to develop more accurate and personalized treatment plans tailored to each patient[41]. For example, generative AI models typically provide similar general clinical advice to cancer patients exhibiting similar signs and symptoms. However, in reality, these patients may have different disease progression and prognoses due to differences in their biomarkers (e.g., DNA, RNA, proteins, metabolites, host cells, microbiomes)[42]. Although collecting and protecting such sensitive data remains a challenge, RAG could better leverage this information for precision medicine practices. Specifically, the RAG system may be able to comprehensively analyze a patient's biomarkers, classify them into more granular subgroups, and recommend appropriate personalized treatment plans to physicians based on established clinical guidelines.

## Conclusion

In conclusion, RAG may enable better integration of generative AI into healthcare and bring more innovative applications in consulting, diagnosis, treatment, management, and education. Despite the potential of RAG systems in medicine, they also face significant limitations. Firstly, the retrieval of external knowledge may introduce additional biases, since the sources themselves might contain biases. Furthermore, due to the lack of sufficient high-quality information on underrepresented groups, RAG systems may become less effective in such cases, with the generated content relying more on the knowledge of models themselves. As a result, minority groups may not benefit much from existing RAG systems. Furthermore, although RAG systems can enhance transparency by providing evidence, it is difficult to know which parts of the response are derived from which pieces of the retrieved knowledge without manual inspection, and even then, it is challenging to understand how the final response is generated. Therefore, we suggest that clinicians, researchers, stakeholders, and regulators collaborate to explore how RAG could be used more equitably, reliably, and effectively to address existing issues in medicine and healthcare.

**Declaration of interests**

This work was supported by the Duke-NUS Signature Research Programme funded by the Ministry of Health, Singapore. Any opinions, findings and conclusions or recommendations expressed in this material are those of the author(s) and do not reflect the views of the Ministry of Health.

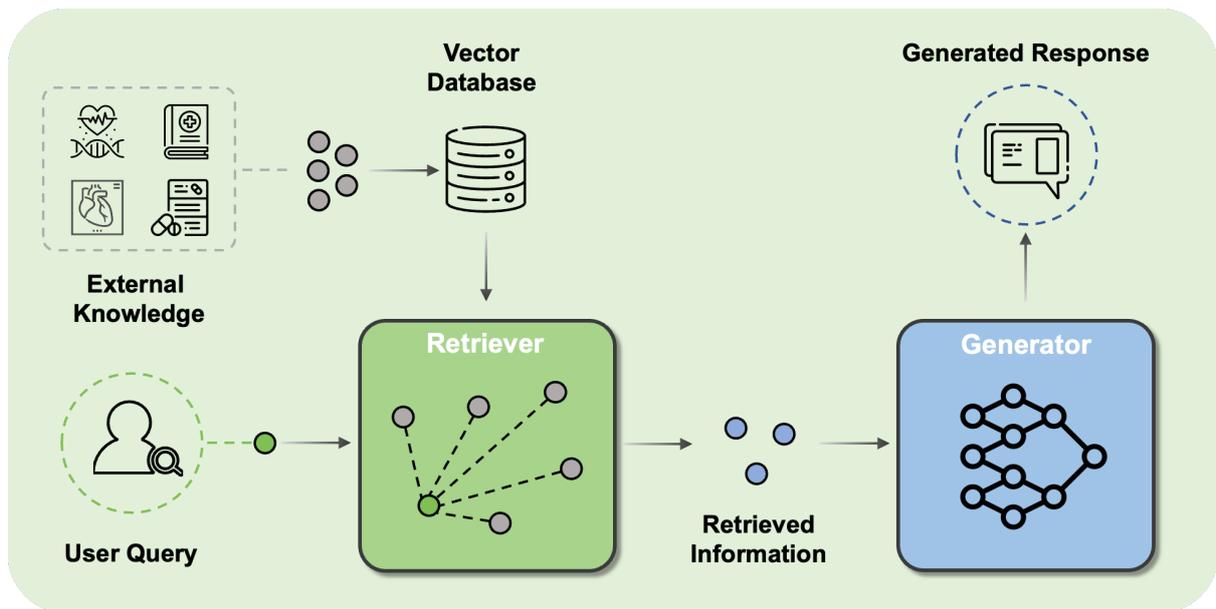

**Figure 1.** A typical retrieval-augmented generation framework. External data is first encoded into vectors and stored in the vector database (where vectors are mathematical representations of various types of data in a high-dimensional space). In the retrieval stage, when receiving a user query, the retriever searches for the most relevant information from the vector database. In the generation stage, both the user's query and the retrieved information are used to prompt the model to generate content.

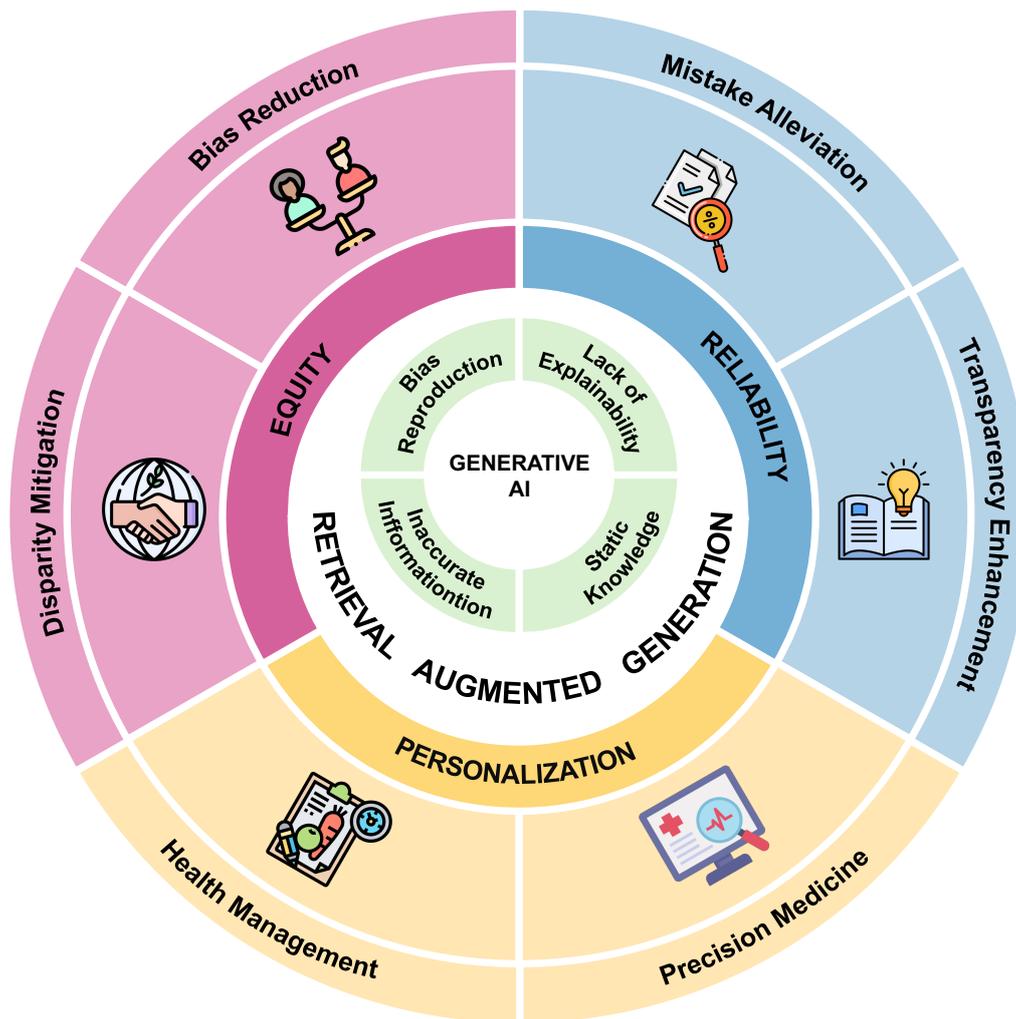

**Figure 2.** Generative AI has limitations such as bias reproduction; lack of transparency; inaccurate information and static knowledge, which hinder its further application in medicine. Retrieval-augmented generation holds promise in alleviating these issues and driving medical innovation in the perspectives of equity, reliability, and personalization.